\documentclass[10pt,twocolumn,letterpaper]{article}

\usepackage{cvpr}
\usepackage{times}
\usepackage{epsfig}
\usepackage{graphicx}
\usepackage{amsmath}
\usepackage{amssymb}
\usepackage{multirow}
\usepackage{array}
\usepackage[noadjust]{cite}
\usepackage{color}
\usepackage{breqn}
\usepackage[british]{babel}
\usepackage{hhline}
\usepackage{tabularx,ragged2e,booktabs,caption}

\usepackage[breaklinks=true,bookmarks=false]{hyperref}

\cvprfinalcopy 

\def\cvprPaperID{****} 

\begin{document}

\title{A Review of Network Inference Techniques for Neural Activation Time Series}

\author{George Panagopoulos\\
{\tt\small Computational Physiology Lab}\\
{\tt\small University of Houston, Houston, TX USA}\\
{\tt\small giorgospanagopoulos.github.io}
}

\maketitle

\begin{abstract}
Studying neural connectivity is considered one of the most promising and challenging areas of modern neuroscience. The underpinnings of cognition are hidden in the way neurons interact with each other. However, our experimental methods of studying real neural connections at a microscopic level are still arduous and costly. An efficient alternative is to infer connectivity based on the neuronal activations using computational methods. A reliable method for network inference, would not only facilitate research of neural circuits without the need of laborious experiments but also reveal insights on the underlying mechanisms of the brain. In this work, we perform a review of methods for neural circuit inference given the activation time series of the neural population. 
Approaching it from machine learning perspective, we divide the methodologies into unsupervised and supervised learning. 
The methods are based on correlation metrics, probabilistic point processes, and neural networks. Furthermore, we add a data mining methodology inspired by influence estimation in social networks as a new supervised learning approach. For comparison, we use the small version of the Chalearn Connectomics competition, that is accompanied with ground truth connections between neurons. The experiments indicate that unsupervised learning methods perform better, however, supervised methods could surpass them given enough data and resources. 
\end{abstract}

\section{Introduction}
With the advent of modern high-throughput neural imaging techniques, neuroscientists have the opportunity to examine individual and collective behaviors of neurons from several organisms in a highly accurate manner \cite{grewe2010high}.
These techniques produce data that contains tremendous insights on how neurons act and collaborate and thus its study may expand our understanding of brain's encoding and decoding processes. However, extracting knowledge from such data is a rather arduous task, due to the complexity , which is  inherent in neural processes, posing numerous methodological challenges. One of these challenges lies in uncovering and analyzing the underlying connectivity of a neural population given their activation patterns, a study that could be placed in the broader field of connectomics \cite{van2010intrinsic}.\\
Examining neural connectivity patterns has proven extremely valuable in multiple levels and modalities. Differences in connections between human brain regions derived from fMRI data, have been indicative of sex \cite{ingalhalikar2014sex},resting state \cite{greicius2003functional}, bipolar disorder \cite{rich2008neural}, mild cognitive impairment \cite{wee2012identification} and creativity \cite{koutra2013deltacon}. Connectivity analysis in EEG data have produced promising results in studying autism \cite{peters2013brain} and epilepsy \cite{bettus2008enhanced}. In the microscopic level, networks of neural cells and synaptic plasticity are widely believed to hold the key to deciphering learning and computation \cite{dayan2003theoretical}. Research on neural circuits though carries additional hindrances, the first one being the lack of experiments with human neural cells, leading to experiments with other organisms such as fish, mice, monkeys or flies. The second is the increasing time and space resolution, compared with other neural data modalities, rendering numerous methods applied in the aforementioned context computationally infeasible or plain simplistic. Assuming the data is acquired in a reliable manner, the neuronal network analysis can be decomposed into two parts. The first is how to define the networks from recordings of neural activations and the second how to mine knowledge from them. The present work addresses the former.\\
One of the most common ways of recording activity in the neural cell level is based on calcium imaging \cite{tian2009imaging}. These images can be converted into neural activation time series and subsequently into spike trains (series of binary events) using algorithms such as Optimal Optical Spike Inference \cite{vogelstein2009oopsi}. Inferring networks from spike trains can be seen as a way to link structural with functional connectivity \cite{guyon2014design}. We focus on three different types of network inference methods, already applied in the literature of connectomics. The first is a family of model-free methods, where the probability of two neurons being connected is equal to a measure of correlation between their time series. 
The second method is an unsupervised probabilistic learning technique, which models the set of spike trains as a Hawkes process. The probability of a spike depends on the previous spikes of other neurons and is conditioned on the parameters that capture their connectivity.
From the point of supervised learning, we examine the use of a residual convolutional neural network, that classifies the connectivity of two neurons given a subset of their activation time series, and ground truth labels. As an extra supervised learning approach, we devise a simple, data mining algorithm that extracts features from the interactions between two neurons and uses an SVM to classify them. We evaluate the methods on a subset of ChaLearn\footnote{http://connectomics.chalearn.org/home} Connectomics challenge dataset. Unsupervised learning is evaluated with all the data, while supervised goes through a leave-one-network-out cross-validation. Unsupervised learning based on partial correlation and the neural network, achieve the highest results, though the former much more efficiently than the latter. However, the neural network could potentially surpass unsupervised approaches, if it is supported by enough data and suitable computational resources. The results also highlight the differences between the predictive behaviors of unsupervised and supervised learning and the advantage of sparse solutions.
The paper is organized as follows. Section 2 describes the different methods employed to perform neural activation discretization and network inference. Section 3 outlines the attributes of the comparative experiment, including the characteristics of the dataset, the evaluation procedure, and the results. Finally, section 4 concludes the paper and points to potentially meaningful directions.


\section{Methods}
\subsection{Preprocessing}
The neural activation series are initially processed to remove the light scattering effect due to fluorescent imaging and subsequently transformed into spike trains based on two different methodologies. The light scattering effect is alleviated by multiplying the activity of all neurons in a given time step with the inverse of a radial distance matrix, computed based on the real distances between the neurons:
\begin{equation}
D_{i,j} = 0.15e^{\frac{|p_i-p_j|^2}{2}}
\end{equation}
In this manner, the effect of light from nearby neurons to the neuron examined is mitigated proportionally to the respective distances between them. \\
Subsequently, we employed the OASIS (Online Active Set method to Infer Spikes)\cite{friedrich2016fast} algorithm, to transform the activations into spike trains.
Finally, a threshold of 0.12 is used to make the series binary, as suggested by the creators of the dataset. Overall, the preprocessing takes on average 30 seconds for each network.\\

\subsection{Unsupervised Learning}
\subsubsection{Model Free Approaches}
Network inference from time series is a task inherently coupled with correlation.
While the mere correlation of activity in a given time interval of approximate neurons might indicate a potential structural connection \cite{bullmore2009complex}, computing an effective connection at a given time relies heavily on time precedence and conditional dependence. A well-known causality metric for such cases is Transfer entropy\cite{schreiber2000measuring}. However, in our early experiments, it significantly underperformed (close to 60\%) and was too computationally expensive to continue using it. Since we can not use a metric that combines both aforementioned properties, we employ two different correlation types. Cross-correlation, which accounts for time precedence and partial correlation, which accounts for conditional dependence. This gives us a chance to also evaluate which of the two properties has a bigger impact on the task at hand.

\paragraph{Cross-Correlation}
Cross-correlation is a standard metric in time series analysis \cite{sarwate1980crosscorrelation}. It captures the correlation between two random variables with a given lag $\tau$. 
\begin{equation}
\gamma_{xy}(\tau) = \sum_t[(x_t-\overline{x})(y_{t+\tau}-\overline{y})]
\end{equation}
The final index is computed by the average cross correlation for all lag values $\tau \in [0,T]$, where $T$ is the total length of the time series. However, due to computational constraints, in our experiments we kept lag = 1. Moreover, to account for directed connections, we compute $\gamma_{yx}$ and $\gamma_{xy}$.

\paragraph{Partial Correlation}
Partial correlation can be measured by the precision matrix, which is the inverse of the covariance $\Sigma^{-1}$, and each element is a measure of conditional independence for the two respective variables. In other words, the element $e_{ij}$ is zero if variables $i$ and $j$ are independent given the rest of the variables in the dataset. That said, accounting for unequivocal causal independence is practically impossible, not only due to computational demand of testing all possible combinations in a given neuronal circuit but also because of the limitations of the imaging method, which may overlook several neurons.\\
Calculating connectivity using an estimate of the precision matrix is quite popular in functional networks of fMRI or EEG data \cite{das2017interpretation}. The nodes in these cases are regions of interest or electrodes. Structural learning models constitute the most dominant approach, mainly due to their efficiency and flexibility, which allows tailoring the model using prior knowledge from neuroscience\cite{craddock2013imaging}. For example, to capture the physiological fact that effective connectivity cannot vary greatly, temporal smoothing regularizations have been utilized \cite{monti2014estimating}. Furthermore, group sparsity is introduced to the model to take advantage of inter-subject similarity for better estimation of each subject's connectivity \cite{varoquaux2010brain}.\\
In our case, we utilize two different approaches to compute partial correlation. The first one is based on the winning solution \cite{sutera2015simple} and essentially estimates the precision matrix through principal component analysis, while keeping the eigenvalues that correspond to roughly 80\% of the variation. The second approach is a straightforward graphical lasso\cite{friedman2008sparse}, as implemented in scikit-learn \cite{scikit-learn}.
\begin{equation}
\Theta' = argmax_{\Theta\leq0}(logdet(\Theta)-tr(\Sigma\Theta)-\lambda|\Theta|_1)
\end{equation}
Since the precision matrix contains negative values, we process it further to match the ground truth connection matrix which is binary. We derive the negative precision matrix, set all diagonal elements to zero, and perform a min-max normalization, which results in a positive matrix with elements in $[0,1]$.

\subsubsection{Hawkes Processes}
Probabilistic approaches are prevalent in neural encoding and decoding literature \cite{paninski2004maximum} because of their ability to capture uncertainty and hierarchy, which agrees with the general Bayesian brain hypothesis \cite{knill2004bayesian}. In addition, they allow for the inclusion of flexible prior distributions that can act as a means to incorporate neuroscientific knowledge in the model, such as network sparseness and smoothness in time\cite{stevenson2009bayesian}.
In this case, we are going to employ the model analyzed by Linderman et al.\cite{linderman2014discovering}, where the activity of a neuron $k$, which is essentially a series of $N$ spikes $s_n \in S$, can be modeled by a conditionally inhomogeneous Poisson process with background rate $\lambda_k(s)$. 
The likelihood of a given set of spikes from neuron $k$ depends on the background rate and the rest events
\begin{dmath}
p({s_n}_{n=1}^N|\lambda_k(s))=e^{-\int_S \lambda_k(s) ds }\prod_{n=1}^N\lambda_k(s_n)
\label{eq:poisson_lik}
\end{dmath}
To include the interactions between neurons, the model has to be expanded to take into account the spike history of all neurons. A Hawkes process is based on events that each belongs to different point processes. Thus the series of spikes $s_n$ is accompanied with a series of labels $c_n \in K$ indicating which of the $K$ processes produced the $n$ spike. By the Poisson superposition theorem \cite{kingman1993poisson} each of these processes can be considered independent, thus the likelihood of a given set of spikes is given by:
\begin{dmath}
p({(s_n,c_n,z_n)}_{n=1}^N|{\lambda_k(s)},{h_{k',k}(\Delta t)}) =\\
\prod_{k=1}^K p({c_n=k, z_n=0}|\lambda_k(s))\\ \times\prod_{n'=1}^N\prod_{k'=1}^K p({c_{n'}=k',z_n=n'}|h_{c_{n'},k}(\Delta t_{n',n}))   	
\label{eq:hawkes_lik}
\end{dmath}
for a spike $s_n$, of neuron $k_n$. $z_n$ is an auxiliary variable added to signify which spike caused spike $n$. 
Each probability in \ref{eq:hawkes_lik} is a Poisson distribution similar to \ref{eq:poisson_lik}. The first component in \ref{eq:hawkes_lik} is a product depicting the probability of $s_n$ being caused by the background firing rate of neuron $k$, in which case $z_n$ is 0 because the spike is not caused by another spike.
The second component represents the probability of spike $n'$ from neuron $k'$ causing spike $n$ of neuron $k$. 
Matrix $h$ captures several aspects of the connectivity between neurons. It can be decomposed into 
\begin{equation}
h_{k',k}(\Delta t) = A_{k',k}W_{k',k}g_{\theta_{k',k}}(\Delta t)
\end{equation}
where $A$ is the adjacency matrix and $W$ is the weight matrix of the network. $g_{\theta_{k',k}}$ is a function parameterized by matrix $\theta$, that captures the time decay between the two neurons and has as input the time passed between the two spikes $\Delta t_{n',n}$.
All the aforementioned parameters are estimated via stochastic variational inference \cite{linderman2015scalable} with 100 iterations. For implementation, we utilized the open source python library pyhawkes. \footnote{https://github.com/slinderman/pyhawkes}

\subsection{Supervised Learning}
\subsubsection{Residual Convolutional Neural Network}
The use of neural networks in the field of connectomics is generally limited, compared to the immense success of deep learning in other fields. Neural networks have been utilized so far in connectomics for image processing tasks, such as identifying neurons in 3D brain imaging \cite{lee2015recursive} or segmentation \cite{turaga2010convolutional}. However, there is not extensive literature based on neural networks for network inference and analysis, with small exceptions, such as a convolutional neural network that identifies contrastive underlying weighted network structures in fMRI data of different subject groups \cite{lee2017identifying}. In this work, we are going to examine a residual convolutional neural network, which is based on the method that achieved fourth place in the competition\cite{romaszko2015signal} and has been extended to a state of the art solution \cite{dunn2017inferring}.\\
This method is accompanied with its own preprocessing, as the model works directly with the fluorescence signals. The signals are downsampled to certain period that is characterized by high overall network activity, meaning that the neurons average  activations surpass 0.02. Afterward, each downsampled signal is standardized, subtracting its mean and dividing by its standard deviation. Subsequently, the data is transformed to serve as input to the neural network. Each training sample consists of a $4x320$ matrix. For each pair of neurons, a subset of 320 is taken from a random starting point of their activation times series.
The first two rows of the training sample correspond to the activity of the two neurons at that subset, the third is the average activity of the whole network and the fourth is the partial correlation between the two neurons, repeated 320 times. The partial correlation is computed after applying a summation filter with length 3 to the two-time series, deriving precision through principal component analysis with 80\% of variance retained similarly to the winner's solution\cite{sutera2015simple}, and standardizing it. To ensure a balance between positive and negative samples in the training dataset, subsets of activity from pairs of connected neurons are sampled more times than the ones with no connection.\\
The architecture of the residual neural network employed\cite{dunn2017inferring} can be seen in figure \ref{fig:rcnn_arch} and in detail consists of:
\begin{figure*}[t]
\centering
\includegraphics[width=1\textwidth]{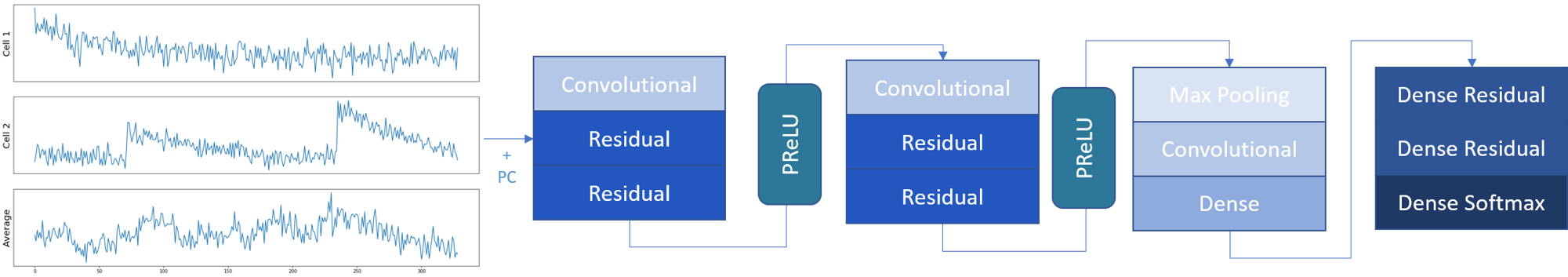}
\caption{Architecture of the Residual Convolutional Neural Network}
\label{fig:rcnn_arch}
\end{figure*}
One convolutional and two residual convolutional layers, each with size [2x326x32] and filters of size [2x5x1]. Another block of one convolutional and two residual convolutional layers, each with size [2x322x64] and filters of size [2x5x32]. A max pooling layer [1x10] and another convolutional layer of size [1x32x128] with filters [1x1x64]. One dense and two residual dense layers with a size of 256 and a dense output layer for softmax classification. The nonlinear function employed is parametric ReLU. The training of the model was performed with Adam optimization, on 100 epochs, a batch size of 100, early stopping in 20 epochs, a dropout probability of 0.2 for convolutional layers and 0.5 for dense layers.
We base our implementation on the python libraries TFconnect \footnote{https://github.com/spoonsso/TFconnect} and TFomics \footnote{https://github.com/spoonsso/tfomics}.
Since this method is supervised, instead of computing the evaluation metrics as done for the model-free methods, we use a leave one network out cross-validation, for all 6 networks. In addition, to get a sense of how the model performs without the partial correlation, which is a standalone solution, we run experiments where the input to the model does not include the partial correlation, and we found out that for this dataset, the two models perform the same.

\subsubsection{Social Influence Model}
As a final method, we have developed a data mining algorithm based on methodologies that model influence dynamics in social networks. The main reason we chose to go this way, is that the problem of estimating who influences whom in a social network, just by observing the time of events, is quite similar to network inference given spike trains. Thus we wanted to examine a potential overlap between these two distinct fields. In addition, the method constitutes another supervised learning approach which contributes to forming a more thorough opinion on whether supervised or unsupervised learning is a better fit for the problem. \\
The idea is basically to extract features from each (directed) pair of neurons, based on the way the spikes of one neuron follow the spikes of the other. 
The steps of the algorithm can be summarized as:
\begin{enumerate}
\itemsep-0.3em
\item Remove samples where over 70\% of the network spikes.
\item Define as candidate impulse responses of a neuron's spike $s$, the immediate next spike $s_n$ of each other neuron.
\item Remove candidate impulse responses that are less than 1 sec away from a spike preceding $s$.
\item For each neuron pair, calculate the time spans between all spikes and the respective impulse responses (e.g. copying time).
\item Transform the time span series into scores using $e^{1.0/(x)}-1$.
\item Compute the number of impulses, mean, variance and 95th percentile of the time span series.
\item Use the features from step 6 as input to an SVM with RBF kernel and adjusted class weights.
\end{enumerate}
The method starts with a preprocessing step. We remove cases where the majority of the network is active simultaneously because the activations might be too noisy due to light scattering.
In the second step, we hypothesize that a spike can potentially cause all the immediate next spikes of the network. Thus each spike $s$ has a pool of $N-1$ possible impulse responses, where $N$ is the number of nodes in the network. 
To reduce this pool, we have to take into account possible fake links that could be derived. For example, if $s$ has two candidate impulse responses $s_1$ at time $t1$, and $s_2$ at $t2>t1$, then $s2$ will be erroneously attributed as a candidate impulse response to $s1$ as well. To refrain from this, when we examine the candidate impulse responses of $s1$, we remove spikes that are less than a second away from a spike preceding $s1$. This is step 2. Of course, it does not fully alleviate the problem, as valid connections could be lost by it and it is heavily dependent on the hyperparameter of 1 second. However, it does improve significantly the result of the algorithm. Subsequently, we calculate all time intervals between spikes and their impulse responses. This gives essentially a series of time spans for each directed pair. In step 4 we transform these series using an exponential function to reward spikes that are very close in time and penalize the opposite. Subsequently, we run simple feature extraction on each series, to end up with a vector of 4 elements that correspond to the training sample for a directed pair.

\section{Experiments}
\subsection{Dataset}
The dataset we use for evaluation stems from an experiment with a larval zebrafish. The zebrafish calcium fluorescence images were used to create a realistic simulator that takes into account the limitations of the imaging technique as well as the real attributes of spiking behavior and neural cultures. In this manner, the calcium fluorescence time series simulated are accompanied with ground truth synaptic connections, in order to evaluate quantitatively the predictive ability of the models.\\
A total of 1000 neurons were simulated together with their activation time series, in a one-hour simulation with the 50Hz sampling frequency, resulting in 180000 samples for each series. The dataset includes 3 different simulated datasets that share the same attributes as the testing dataset. Due to computational constraints, we limit our experiments in 6 small networks, with 100 neurons each, provided by the organizers of the competition. 

\subsection{Evaluation}
The number of real connections in each network corresponds to only ~10\% of the overall possible connections, which means that accuracy is not a completely reliable method for validation. That is why we employ Area Under Receiver Operator Curve (AUC) and precision Recall Curve (PRC) for evaluation. The former is one of the most prevalent evaluation metrics in machine learning literature and intuitively captures the relationship between sensitivity and specificity, with 100\% being the result of an optimum classifier. 
The latter captures the relationship between high precision and low recall when it is low, and it has proven useful for network inference in bioinformatics' low-density networks \cite{schrynemackers2013protocols}. As mentioned above, unsupervised learning is evaluated with all the data, while supervised goes through a leave-one-network-out cross-validation, meaning that we train on five networks and test on the sixth in a repeated manner. The code to reproduce the experiments can be found on github\footnote{https://github.com/GiorgosPanagopoulos/Network-Inference-From-Neural-Activations}.

\begin{table*}[!t]
  \centering
\begin{tabular}{ | c | c | c | c |}
\hline
\textbf{Network Inference Method} & \textbf{AUC \%} & \textbf{PRC \%}& \textbf{Time (Sec)}\\
\hline
Graphical Lasso &83.1 & 44.2 & 40 \\
 \hline
RCNN & 83 & 44.9 & 6082\\
\hline
Cross Correlation &77.7 & 34.9 & 432 \\
 \hline
PCA & 76.1 & 30.7 & 33  \\
 \hline
Hawkes & 72.8 & 35.5 & 5588\\
 \hline
 CIRUSIM & 68.7 & 19.3 & 3276\\
 \hline
\end{tabular}
\captionof{table}{Area under the ROC Curve (AUC) and Precision Recall Curve (PRC). Computation time accounts for one run of a cross validation fold.} 
\label{tab:results_table}
\end{table*}

\begin{figure*}[t]
\centering
\includegraphics[width=1\textwidth]{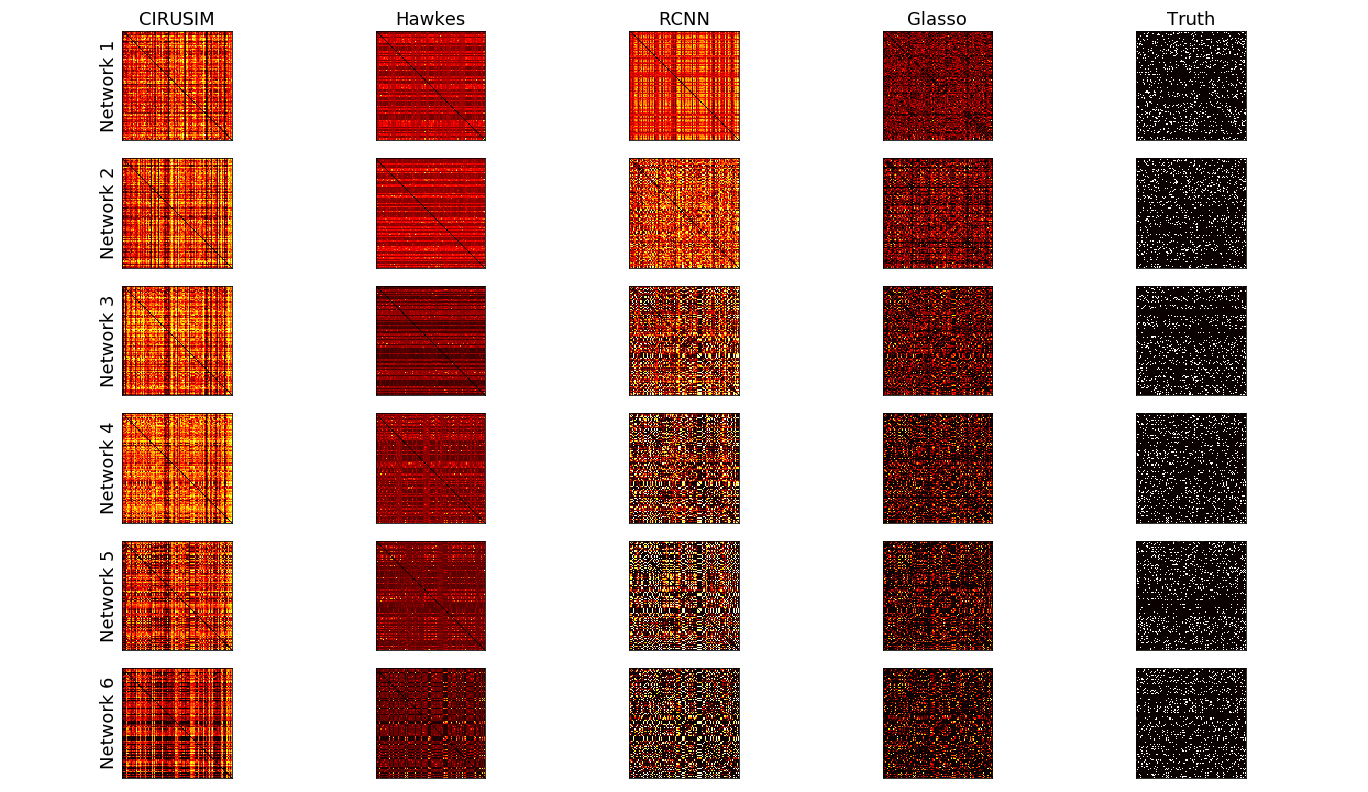}
\caption{Heatmaps of the inferred connectivity matrices and the ground truth network. The lighter the color, the higher the probability of connection.}
\label{fig:connectivities}
\end{figure*}
\subsection{Results}

Table \ref{tab:results_table} shows the cross-validation AUC and PRC for each method as well as the average time needed to train and test one cross-validation fold. 
Graphical lasso clearly outperforms the rest of the methods in terms of accuracy and is the second fastest. This comes in accordance with the literature of connectomics where partial correlation is extensively employed, as mentioned above. 
We hypothesize that the reason behind graphical lasso outperforming the winning solution (PCA) was due graphical lasso's sparsity constraint. More specifically, lasso facilitates retrieving sparse connections, which is a prime characteristic of the networks we try to infer.
Similar examples where sparsity regularizations on the precision matrix have enhanced identification of neural connections exist in the literature \cite{huang2009learning}.
Cross-correlation comes second in accuracy for the unsupervised learning methods, though it is much slower then PCA and graphical lasso. Possibly if the lag hyperparameter is optimized, cross-correlation could get even higher. Hawkes process performs quite purely. One possible explanation could be the limited number of iterations (100) which must have a strong negative impact on stochastic variational inference. Moving to the supervised models, RCNN had almost the same accuracy as the graphical lasso, but by far the worst computation time. The model would surely achieve extreme acceleration if it is run on a GPU. Moreover, this algorithm was developed and evaluated in networks of 1000 neurons, which means the number of training samples used to train it is orders of magnitude bigger. Given the need of deep learning techniques for voluminous data, the accuracy of the model was unexpectedly high. Finally, the model based on social influence (CIRUSIM) has the worst accuracy and is overly time inefficient. This could be due to the simplistic basis of the method or the several hyperparameters that need tuning. However, CIRUSIM achieves 81\% AUC with standard 5-fold cross-validation. This could mean supervised learning performs worse then unsupervised because it does not take into account the properties of the test network. \\
To dive deeper into the results, figure \ref{fig:connectivities} shows heat maps with the inferred connectivity probability matrices. We chose graphical lasso to represent the model-free approaches because it is the most successful of them. One can see that supervised learning models tend to produce higher probabilities then unsupervised ones. This is more prevalent in CIRUSIM, but it is also visible in the first 3 networks for RCNN. This predictive behavior is not favored by metrics such as AUC, which penalize harshly false positives. The improvement in performance when the solution gets more sparse or the probabilities fall, is visible in the AUC of these models;  CIRUSIM has an average AUC of 61\% on the first three networks and 75\% in the last three, while RCNN has 74\% on the first three and 91\% on the last. Generally, the last three networks were easier to predict for both, unsupervised and supervised networks. The Hawkes process has an overly sparse solution, where the predicted connections are very few and as a result, the true positives are minuscule. 
In contrast, the solution of graphical lasso is more balanced. It is quite sparse but it has a fair amount of high probabilities. RCNN clearly infers more connections then Glasso. More specifically, it has an average of 10\% higher precision in each network. However, the sparsity of Glasso makes up for it with consistently great recall (over 92\% on average).

\section{Future Work}

Our experiments showed that unsupervised learning provides a more efficient solution. However, we know that RCNN has the strongest solution with networks of 1000 cells \cite{dunn2017inferring} and as we showed, it has a very competitive performance even with networks of 100 cells. Thus, the use of supervised methods can be justified, if certain requirements regarding the computational resources and the volume of data are satisfied. Of course, the existence of ground truth labels is also a prerequisite, probably the hardest one. 
If we could summarize the conclusion of the study in one suggestion, it would be to use graphical lasso when the data is limited and RCNN otherwise. However, to define how much data suffice to employ a neural network approach, several experiments need to take place with a diverse set of data.\\
Given the difference between leave-one-network-out and simple cross-validation, we can assume that the accuracy of supervised methods would be enhanced if they could also learn from the test networks. Hence, examining a semi-supervised learning approach could provide a new perspective and stronger results.
Moreover, the experiments reveal the importance of sparse solutions. This sparsity could be guided based on prior neuroscientific knowledge, in the prototypes of previous studies \cite{varoquaux2010brain,monti2014estimating}. For example, multi-task learning methods have proven promising in alleviating the negative effect of inter-network variations, even with few and noisy data \cite{panagopoulos2017multi}.\\
Finally, in this study, we have overlooked two classes of powerful connectivity inference methodologies. One is based on generalized linear models \cite{linderman2016bayesian} and the other on stochastic leaky integrate-and-fire models \cite{paninski2004maximum}. These two approaches need to be taken into consideration to form a more catholic view of the available solutions. 

\setlength{\textfloatsep}{5pt}


{\small
\bibliographystyle{ieee}
\bibliography{bib}
}

\end{document}